\documentclass{article}

\usepackage[nonatbib]{neurips_2019}
\usepackage[numbers]{natbib}

\usepackage[utf8]{inputenc} 
\usepackage[T1]{fontenc}    
\usepackage{hyperref}       
\usepackage{url}            
\usepackage{booktabs}       
\usepackage{amsfonts}       
\usepackage{nicefrac}       
\usepackage{microtype}      
\usepackage{graphicx}
\usepackage{hyperref}
\usepackage[dvipsnames]{xcolor}
\usepackage[normalem]{ulem}
\usepackage{hyperref}       
\usepackage{url}            
\usepackage{booktabs}       
\usepackage{amsfonts}       
\usepackage{nicefrac}       
\usepackage{microtype}      
\usepackage{multirow}
\usepackage{graphicx}
\usepackage{colortbl}
\usepackage{amsmath}
\usepackage{float}
\usepackage{booktabs}
\usepackage{multirow}
\usepackage{tabularx}
\usepackage{graphics}

\usepackage{pgfplots}
\usepackage{float}

\setcitestyle{square}
\newif{\ifhidecomments}

\usepackage{mathtools}

\usepackage{float}

\usepackage{array}
\newcolumntype{P}[1]{>{\centering\arraybackslash}p{#1}}
\newcolumntype{M}[1]{>{\centering\arraybackslash}m{#1}}

\usepackage{tikz}

\nolinenumbers
\title{[Re] Double Sampling Randomized Smoothing}

\author{
  Aryan Gupta \\
  Department of Electrical Engineering \\
  Indian Institute of Technology, Roorkee \\
  \texttt{a\_gupta@ee.iitr.ac.in} \\
  \And
  Sarthak Gupta \\
  Department of Mathematics \\
  Indian Institute of Technology, Roorkee \\
  \texttt{sarthak\_g@ma.iitr.ac.in} \\
  \And
  Abhay Kumar \\
  Department of Mathematics \\
  Indian Institute of Technology, Roorkee \\
  \texttt{a\_kumar2@ma.iitr.ac.in} \\ 
  \And
  Harsh Dugar \\
  Department of Mathematics \\
  Indian Institute of Technology, Roorkee \\
  \texttt{h\_dugar@ma.iitr.ac.in} \\ 
}

\begin{document}
\maketitle
\section*{\centering Reproducibility Summary}


\subsection*{Scope of Reproducibility}

This effort aims to reproduce and validate the experiments and robustness of the \textit{Double Sampling Randomized Smoothing}\cite{https://doi.org/10.48550/arxiv.2206.07912} certification technique introduced in \cite{https://doi.org/10.48550/arxiv.2206.07912}. We consistently verify the improvement in robustness radius for the DSRS certification compared to the standard Neyman-Pearson Certification\cite{https://doi.org/10.48550/arxiv.2002.08118}\cite{https://doi.org/10.48550/arxiv.1902.02918}\cite{https://doi.org/10.48550/arxiv.1906.04584}\cite{https://doi.org/10.48550/arxiv.1809.03113} as reported in \cite{https://doi.org/10.48550/arxiv.2206.07912}. We also study the effectiveness of the certification through adversarial trained models, ablation studies and new experiments.

\subsection*{Methodology}

We start the reproduction effort by using the code open-sourced by the authors. We reproduce Tables 7 and 8 of the original paper. Further, we refactor their code for evaluating the method's effectiveness under various circumstances. We implement new ablation studies to test the method and design experiments to verify the claims made by the authors. We release our code as open-source.

\subsection*{Results}

We reproduced the results of DSRS certification on Cifar‐10 trained Resnet‐110 model within 1.7\% accuracy of the reported values, and about half times we reproduced even higher accuracy than the reported ones in the paper. The authors' claim that DSRS provides consistently tighter robustness certification than existing baselines, seems to be consistently verified. The ablation studies help us to further strengthen the claims made by the authors and get further insights.

\subsection*{What was easy}

The authors open-sourced the code for the paper. This made it easy to verify the results and allowed us to easily make edits for ablation studies. The mechanism was also mathematically well described in the paper, helping us conceptually understand the methods. 

\subsection*{What was difficult}

The primary difficulty that we faced was the limitation of our computational resources as compared to the requirements for conducting the paper's experiments and further ablation studies. For instance, some of the experiments took over 60 hours to run.

\subsection*{Communication with original authors}
The authors assisted us with key insights in this reproducibility challenge, from providing us the main code of the paper to their supportive suggestions to our ablation studies. We would like to acknowledge Prof. Bo Li, Linyi Li, and others for their invaluable time. Their feedback for this report is available in the summary \ref{AuthorsComment}.
\newpage

\section{Introduction}
Verifying the robustness of deep learning models is an active area of research. There have been several efforts towards this direction \cite{DBLP:journals/corr/abs-1711-00851}\cite{https://doi.org/10.48550/arxiv.1802.03471}. One significant work was \textit{Certified Adversarial Robustness via Randomized Smoothing}\cite{https://doi.org/10.48550/arxiv.1902.02918}. While it provided a method for effectively verifying the robust radius of a deep network, it was not able to scale with the increasing input dimensions. To improve this, \textit{Double Sampling Randomized Smoothing} provides a mathematical framework which builds upon \cite{https://doi.org/10.48550/arxiv.1902.02918}. The paper proposes to use an additional distribution to calculate a tighter robust radius around the samples. This distribution provides additional data for improving the robust radius and can certify radius $\Omega(\sqrt[]{d})$ under the $l_2$  norm.

\section{Scope of reproducibility} \label{scope}
\label{sec:claims}

Under this reproducibility effort, we focus on reproducing the original results of the paper as well as conducting further experiments to validate the mechanism in varying settings. The claims from the original paper, which we attempt to verify, are listed as follows.
\begin{itemize}
    \item DSRS certifies $\Omega(\sqrt[]{d})$ under the $l_2$  norm being the input dimension, under mild assumptions. This implies that DSRS is able to break the curse of dimensionality of randomized smoothing(“curse of dimensionality” or “$l_\infty$ barrier”)\cite{https://doi.org/10.48550/arxiv.2002.08118}\cite{https://doi.org/10.48550/arxiv.2002.03517}\cite{https://doi.org/10.48550/arxiv.2002.03239}\cite{https://doi.org/10.48550/arxiv.2006.04208}.
    \item  DSRS provides consistently tighter robustness certification than existing baselines, including standard Neyman-Pearson based certification across different models on MNIST and CIFAR-10.
\end{itemize}

Further, we try to evaluate the proposed method by conducting the following experiments:
\begin{itemize}
    \item Testing the DSRS method by varying the amount of samples.
    \item Evaluating the effects of adversarial training on the certified radius of the smoothed classifier.
    \item Evaluating different distributions for smoothing than the ones used in the original work.
    \item Observing the variations in variance of the distributions used to smooth the classifier.
\end{itemize}

\section{Methodology}
For this reproducibility study, we begin with the open-sourced code provided by the authors and refactor it to conduct ablations wherever necessary. To verify the claims listed in Section \ref{sec:claims} we make use of the MNIST \cite{deng2012mnist} and CIFAR-10 \cite{cifar_cite}. Initially, we recreate the experiments exactly as listed in \cite{https://doi.org/10.48550/arxiv.2206.07912} and then conduct several ablation studies by varying different hyperparameters and configurations in the proposed method.

\subsection{DSRS Algorithm Description}

\begin{figure}
    \centering
    \includegraphics[width = 0.8\textwidth]{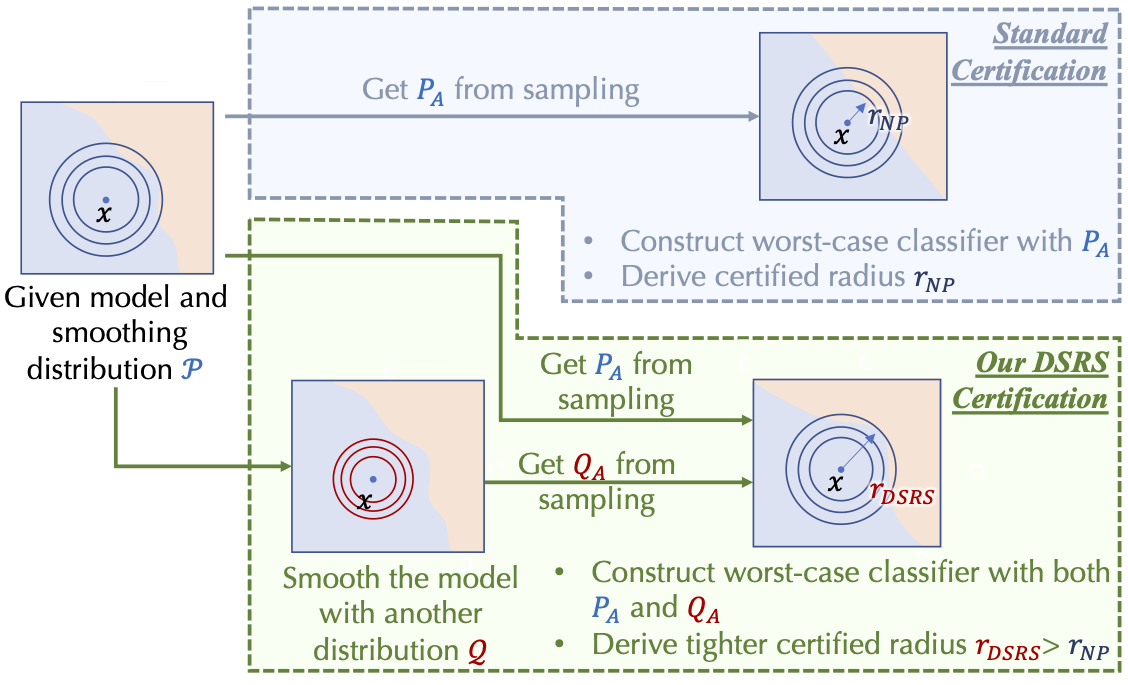}
    \caption{\label{algo} A high-level overview of the DSRS method. \cite{https://doi.org/10.48550/arxiv.2206.07912} }
    \label{fig:dsrs_basic_pipeline}
\end{figure}

Double sampling randomized smoothing (DSRS) is a novel robustness certification framework for randomized smoothing. Randomized smoothing (RS) \cite{https://doi.org/10.48550/arxiv.1902.02918} adds noise to the input and uses majority voting among the model predictions for multiple noised inputs to get the final label prediction. Since the shifting of the center of noised input distribution does not change the ranking of label predictions, the RS models are certifiably robust to small $l_2$-norm-bounded perturbations.

Given an input instance and a model, robustness certification approach computes a robust radius, such that any perturbations within the radius does not the change the final prediction. By nature, certification approach is conservative --- it provides a lower bound of robust radius while the maximum robust radius for the given input instance could be much larger than the one computed by the certification approach.

For RS, the most widely-used certification approach is the Neyman-Pearson-based certification\cite{https://doi.org/10.48550/arxiv.2002.08118}\cite{https://doi.org/10.48550/arxiv.1902.02918}\cite{https://doi.org/10.48550/arxiv.1906.04584}\cite{https://doi.org/10.48550/arxiv.1809.03113}. It leverages the probability of base model's predicting each class under the input noise to compute the certification.

In DSRS, authors propose to sample the base model's prediction statistics under two different distributions, and leverage the joint information to compute the certification. Since more information is being utilized, the certification approach is guaranteed to be at least as tight as the most widely-used Neyman-Pearson-based approach.


\subsection{Hyperparameters}
The main hyperparameters in the certification are related to the two sampling distributions i.e. P and Q. We studied the effect of changing the hyperparameter k of the generalized gaussian distributions\cite{https://doi.org/10.48550/arxiv.2002.09169}. Also we conducted experiments on variances $\sigma_P$ and $\sigma_Q$ of the main distributions and studied the effects of their variation on the results.

\subsection{Experimental setup, Datasets and code}
For our studies, we use the MNIST and CIFAR-10 datasets containing 60,000 and 50,000 train examples respectively. We created bash scripts for each setting of the experiment depending on the datasets and variance of the sampling distributions of P(0.25, 0.50, and 1.00) and Q(0.20, 0.40 and 0.80), and the distributions for sampling noise i.e. Standard Gaussian\cite{https://doi.org/10.48550/arxiv.1902.02918}\cite{https://doi.org/10.48550/arxiv.2002.08118}, and Generalized Gaussian\cite{https://doi.org/10.48550/arxiv.2002.09169}(k=380 for Mnist and k=1530 for Cifar10).Finally we trained each model on three commonly used or state-of-the-art training methods: Gaussian augmentation \cite{https://doi.org/10.48550/arxiv.1902.02918}, Consistency \cite{https://doi.org/10.48550/arxiv.2006.04062}, and SmoothMix \cite{https://doi.org/10.48550/arxiv.2111.09277}. We run the main experiments of the paper 3 times and provide 95\% confidence intervals for the results in the Section \ref{sec:results}. Also we created bash scripts for the respective ablation studies like training adversarial models and hyperparameter search to allow easy reproduction of experiments.

\subsection{Computational requirements}
We use a single NVIDIA Tesla V100 GPU with 32 GBs of RAM for this study. The computational requirements for validating the results and conducting ablations were high when compared to the compute power at our disposal. The certification of robust radius for MNIST took over ten hours in all cases. For CIFAR-10, the certification took over twenty-five hours, with some experiments taking over sixty-five hours.

\section{Results}
\label{sec:results}
In this section, we present the reproduction of certification results and tables associated with the MNIST and CIFAR-10 datasets that the original paper used to support their method. We could not use the larger ImageNet dataset due to limited computational resources. We cover the results for MNIST and CIFAR-10 in the same form as given in the original paper and conduct some ablations to further investigate the DSRS framework.

\subsection{Results reproducing original paper}
Reproducing the original results of the paper helps us understand the main claims of the paper, as mentioned in Section~\ref{sec:claims}, and the main advantages of DSRS certification over Standard Neyman-Pearson certification. 

\subsubsection{Verification of robustness radius having order $\Omega(\sqrt[]{d})$ }\label{maincurve}
The DSRS framework seems to work well on increasing dimensions and provides us with an increased ACR(Average Certified Radius)\cite{https://doi.org/10.48550/arxiv.2001.02378} as compared to the standard Neyman-Pearson certification. Also, we can see the effect of varying sigma values on the ACR values.

\begin{figure}[!ht]
    \centering
    \includegraphics[width = 0.8\textwidth]{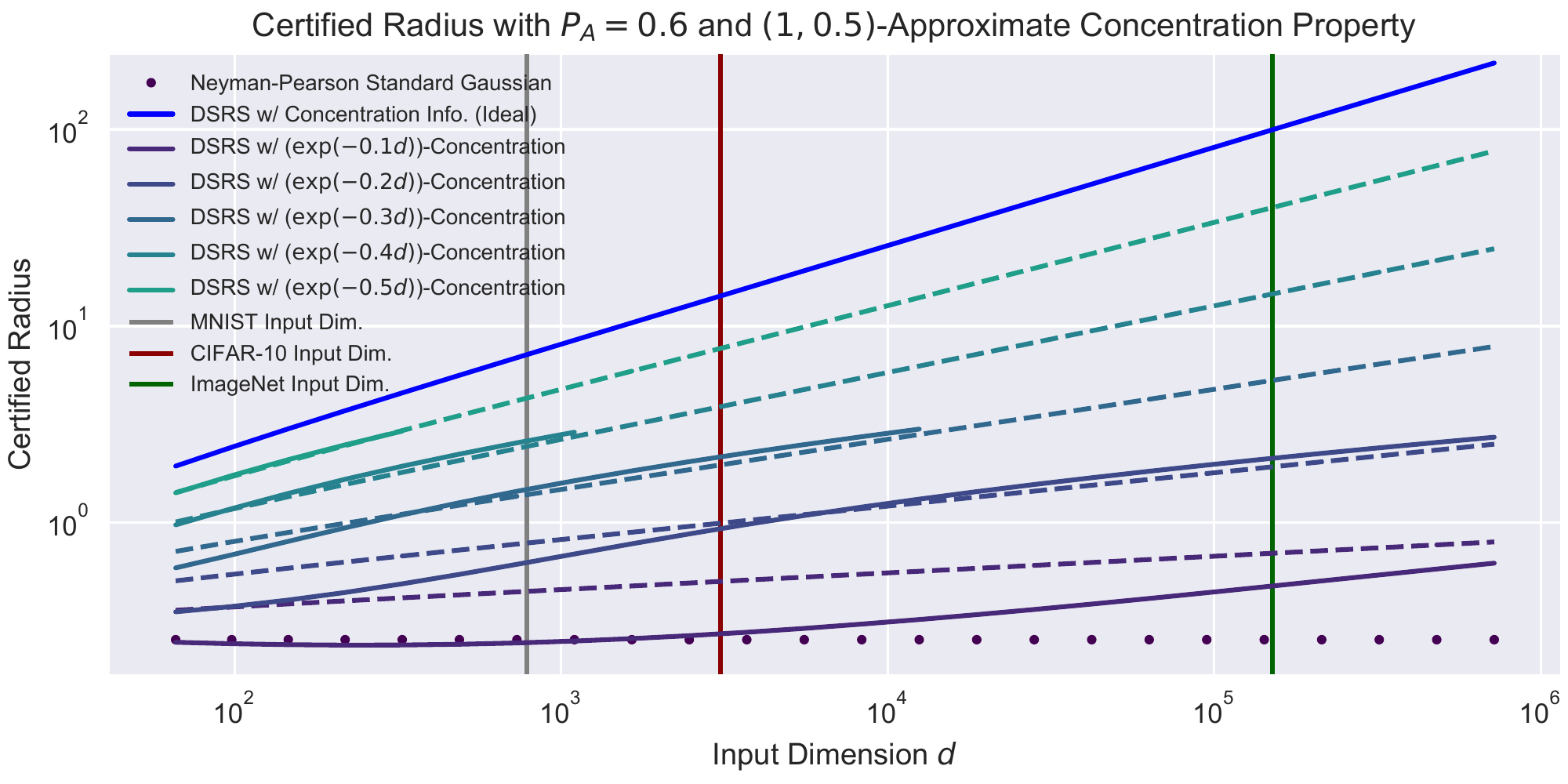}
    \caption{\label{res} The main result of DSRS method.\cite{https://doi.org/10.48550/arxiv.2206.07912} }
    \label{fig:dsrs_main_result}
\end{figure}

\subsubsection{Results verifying higher certified radius for DSRS on MNIST and CIFAR10}
The major results of the main paper are reproduced in this section. We have reproduced these results three times and have calculated the confidence intervals appropriately. Also, insights have been made into the results and their variation from the ones provided in the main paper.

\begin{table}[H]
\centering
\caption{MNIST: General Gaussian with 95\% confidence interval}
\label{table:MNIST Main}
\resizebox{16cm}{!}{%
\begin{tabular}{c|c|c|cccccccccccc}
\hline
\multirow{2}{*}{Variance} & \multirow{2}{*}{\begin{tabular}[c]{@{}c@{}}Training\\  Method\end{tabular}} & \multirow{2}{*}{\begin{tabular}[c]{@{}c@{}}Certification\\  Approach\end{tabular}} & \multicolumn{12}{c}{Certified Accuracy under Radius r} \\
 &  &  & 0.25 & 0.50 & 0.75 & 1.00 & 1.25 & 1.50 & 1.75 & 2.00 & 2.25 & 2.50 & 2.75 & 3.00 \\ \hline
\multirow{9}{*}{0.25} & \multirow{3}{*}{\begin{tabular}[c]{@{}c@{}}Gaussian\\  Augmentation\end{tabular}} & Neyman-Pearson & 97.9\% & 96.4\% & 92.1\% &  &  &  &  &  &  &  &  &  \\
 &  & DSRS (OP) & 97.9\% & 96.6\% & 92.7\% &  &  &  &  &  &  &  &  &  \\
 &  & DSRS(Ours) & \textbf{97.8±0.3\%} & \textbf{96.7±0.1\%\textsuperscript{*}} & \textbf{93.0±0.4\%\textsuperscript{*}} &  &  &  &  &  &  &  &  &  \\ \cline{2-15} 
 & \multirow{3}{*}{Consistency} & Neyman-Pearson & 98.4\% & 97.5\% & 94.4\% &  &  &  &  &  &  &  &  &  \\
 &  & DSRS (OP) & 98.4\% & 97.5\% & 95.4\% &  &  &  &  &  &  &  &  &  \\
 &  & DSRS(Ours) & \textbf{98.3±0.1\%} & \textbf{97.5±0.3\%} & \textbf{95.1±0.3\%} &  &  &  &  &  &  &  &  &  \\ \cline{2-15} 
 & \multirow{3}{*}{SmoothMix} & Neyman-Pearson & 98.6\% & 97.6\% & 96.5\% &  &  &  &  &  &  &  &  &  \\
 &  & DSRS (OP) & 98.6\% & 97.7\% & 96.8\% &  &  &  &  &  &  &  &  &  \\
 &  & DSRS(Ours) & \textbf{98.6±0.2\%} & \textbf{97.2±0.3\%} & \textbf{95.5±0.2\%} &  &  &  &  &  &  &  &  &  \\ \hline
\multirow{9}{*}{0.50} & \multirow{3}{*}{\begin{tabular}[c]{@{}c@{}}Gaussian\\  Augmentation\end{tabular}} & Neyman-Pearson & 97.8\% & 96.9\% & 94.6\% & 88.4\% & 78.7\% & 52.6\% &  &  &  &  &  &  \\
 &  & DSRS (OP) & 97.8\% & 97.0\% & 95.0\% & 89.8\% & 83.4\% & 59.1\% &  &  &  &  &  &  \\
 &  & DSRS(Ours) & \textbf{97.9±0.2\%\textsuperscript{*}} & \textbf{97.0±0.1\%} & \textbf{95.3±0.5\%\textsuperscript{*}} & \textbf{90.5±0.2\%\textsuperscript{*}} & \textbf{82.6±1.0\%} & \textbf{59.8±1.8\%\textsuperscript{*}} &  &  &  &  &  &  \\ \cline{2-15} 
 & \multirow{3}{*}{Consistency} & Neyman-Pearson & 98.4\% & 97.3\% & 96.0\% & 92.3\% & 83.8\% & 67.5\% &  &  &  &  &  &  \\
 &  & DSRS (OP) & 98.4\% & 97.3\% & 96.0\% & 93.5\% & 87.1\% & 71.8\% &  &  &  &  &  &  \\
 &  & DSRS(Ours) & \textbf{98.3±0.3\%} & \textbf{97.3±0.2\%} & \textbf{96.1±0.3\%\textsuperscript{*}} & \textbf{93.4±0.4\%} & \textbf{87.0±0.4\%} & \textbf{71.2±0.8\%} &  &  &  &  &  &  \\ \cline{2-15} 
 & \multirow{3}{*}{SmoothMix} & Neyman-Pearson & 98.2\% & 97.1\% & 95.4\% & 91.9\% & 85.1\% & 73.0\% &  &  &  &  &  &  \\
 &  & DSRS (OP) & 98.1\% & 97.1\% & 95.9\% & 93.4\% & 87.5\% & 76.6\% &  &  &  &  &  &  \\
 &  & DSRS(Ours) & \textbf{97.5±0.1\%} & \textbf{96.4±0.1\%} & \textbf{94.5±0.1\%} & \textbf{91.9±0.2\%} & \multicolumn{1}{l}{\textbf{86.3±0.2\%}} & \textbf{75.4±0.6\%} &  &  &  &  &  &  \\ \hline
\multirow{9}{*}{1.00} & \multirow{3}{*}{\begin{tabular}[c]{@{}c@{}}Gaussian\\  Augmentation\end{tabular}} & Neyman-Pearson & 95.2\% & 91.9\% & 87.7\% & 80.6\% & 71.2\% & 57.6\% & 41.0\% & 25.5\% & 13.6\% & 6.2\% & 2.1\% & 0.9\% \\
 &  & DSRS (OP) & 95.1\% & 91.8\% & 88.2\% & 81.5\% & 73.6\% & 61.6\% & 48.4\% & 34.1\% & 21.0\% & 10.6\% & 4.4\% & 1.2\% \\
 &  & DSRS(Ours) & \textbf{94.8±0.9\%} & \textbf{91.2±0.9\%} & \textbf{86.8±1.7\%} & \textbf{80.3±1.6\%} & \textbf{72.5±3.3\%} & \textbf{60.6±2.6\%} & \textbf{47.9±1.8\%} & \textbf{32.7±2.2\%} & \textbf{19.6±2.7\%} & \textbf{9.7±1.5\%} & \textbf{4.4±0.2\%} & \textbf{1.1±0.2\%} \\ \cline{2-15} 
 & \multirow{3}{*}{Consistency} & Neyman-Pearson & 93.9\% & 90.9\% & 86.4\% & 80.8\% & 73.0\% & 61.1\% & 49.1\% & 35.6\% & 21.7\% & 10.4\% & 4.1\% & 1.9\% \\
 &  & DSRS (OP) & 93.9\% & 91.1\% & 86.9\% & 81.7\% & 75.2\% & 65.6\% & 55.8\% & 41.9\% & 31.4\% & 17.8\% & 8.6\% & 2.8\% \\
 &  & DSRS(Ours) & \textbf{94.2±0.2\%\textsuperscript{*}} & \textbf{91.2±0.1\%\textsuperscript{*}} & \textbf{87.3±0.2\%\textsuperscript{*}} & \textbf{81.7±0.3\%} & \textbf{74.7±0.2\%} & \textbf{65.6±0.7\%} & \textbf{54.7±1.0\%} & \textbf{41.9±0.8\%} & \textbf{30.2±1.1\%} & \textbf{17.1±1.5\%} & \textbf{8.6±0.1\%} & \textbf{2.6±0.1\%} \\ \cline{2-15} 
 & \multirow{3}{*}{SmoothMix} & Neyman-Pearson & 92.0\% & 88.9\% & 84.4\% & 78.6\% & 69.8\% & 60.7\% & 49.9\% & 40.2\% & 31.5\% & 22.2\% & 12.2\% & 4.9\% \\
 &  & DSRS (OP) & 92.2\% & 89.0\% & 84.8\% & 79.7\% & 72.0\% & 63.9\% & 54.4\% & 46.2\% & 37.6\% & 29.2\% & 18.5\% & 7.2\% \\
 &  & DSRS(Ours) & \textbf{91.9±0.1\%} & \textbf{88.7±0.2\%} & \textbf{85.0±0.1\%} & \textbf{79.8±0.3\%\textsuperscript{*}} & \textbf{72.4±0.2\%\textsuperscript{*}} & \textbf{63.9±0.1\%} & \textbf{54.9±0.2\%\textsuperscript{*}} & \textbf{46.1±0.3\%} & \textbf{37.5±0.1\%} & \textbf{29.0±0.2\%} & \textbf{19.1±0.3\%\textsuperscript{*}} & \textbf{7.2±0.2\%} \\ \hline
\end{tabular}%
}
\end{table}

\begin{table}[H]
\centering
\caption{CIFAR-10:
General Gaussian with 95\% confidence interval}
\label{table:CIFAR-10 Main}
\resizebox{16cm}{!}{%
\begin{tabular}{c|c|c|cccccccccccc}
\hline
\multirow{2}{*}{Variance} & \multirow{2}{*}{\begin{tabular}[c]{@{}c@{}}Training\\  Method\end{tabular}} & \multirow{2}{*}{\begin{tabular}[c]{@{}c@{}}Certification\\  Approach\end{tabular}} & \multicolumn{12}{c}{Certified Accuracy under Radius r} \\
 &  &  & 0.25 & 0.50 & 0.75 & 1.00 & 1.25 & 1.50 & 1.75 & 2.00 & 2.25 & 2.50 & 2.75 & 3.00 \\ \hline
\multirow{9}{*}{0.25} & \multirow{3}{*}{\begin{tabular}[c]{@{}c@{}}Gaussian\\  Augmentation\end{tabular}} & Neyman-Pearson & 56.1\% & 35.7\% & 13.4\% &  &  &  &  &  &  &  &  &  \\
 &  & DSRS (OP) & 57.4\% & 39.4\% & 17.3\% &  &  &  &  &  &  &  &  &  \\
 &  & DSRS(Ours) & \textbf{57.7±0.8\%\textsuperscript{*}} & \textbf{37.8±0.7\%} & \textbf{17.5±0.4\%\textsuperscript{*}} &  &  &  &  &  &  &  &  &  \\ \cline{2-15} 
 & \multirow{3}{*}{Consistency} & Neyman-Pearson & 61.8\% & 50.9\% & 34.7\% &  &  &  &  &  &  &  &  &  \\
 &  & DSRS (OP) & 62.5\% & 52.5\% & 37.8\% &  &  &  &  &  &  &  &  &  \\
 &  & DSRS(Ours) & \textbf{62.8±0.7\%\textsuperscript{*}} & \textbf{52.9±0.8\%\textsuperscript{*}} & \textbf{39.2±0.9\%\textsuperscript{*}} &  &  &  &  &  &  &  &  &  \\ \cline{2-15} 
 & \multirow{3}{*}{SmoothMix} & Neyman-Pearson & 63.9\% & 53.3\% & 38.4\% &  &  &  &  &  &  &  &  &  \\
 &  & DSRS (OP) & 64.7\% & 55.5\% & 41.1\% &  &  &  &  &  &  &  &  &  \\
 &  & DSRS(Ours) & \textbf{60.5±0.6\%} & \multicolumn{1}{l}{\textbf{53.2±0.6\%}} & \multicolumn{1}{l}{\textbf{42.5±0.5\%\textsuperscript{*}}} &  &  &  &  &  &  &  &  &  \\ \hline
\multirow{9}{*}{0.50} & \multirow{3}{*}{\begin{tabular}[c]{@{}c@{}}Gaussian\\  Augmentation\end{tabular}} & Neyman-Pearson & 53.7\% & 41.3\% & 27.7\% & 17.1\% & 9.1\% & 2.8\% &  &  &  &  &  &  \\
 &  & DSRS (OP) & 54.1\% & 42.7\% & 30.6\% & 20.3\% & 12.6\% & 4.0\% &  &  &  &  &  &  \\
 &  & DSRS(Ours) & \textbf{52.8±0.8\%} & \textbf{40.6±1.4\%} & \textbf{29.6±1.5\%} & \multicolumn{1}{l}{\textbf{20.2±0.9\%}} & \textbf{12.0±0.3\%} & \multicolumn{1}{l}{\textbf{4.7±0.5\%}\textsuperscript{*}} &  &  &  &  &  &  \\ \cline{2-15} 
 & \multirow{3}{*}{Consistency} & Neyman-Pearson & 49.2\% & 43.9\% & 38.0\% & 32.3\% & 23.8\% & 18.1\% &  &  &  &  &  &  \\
 &  & DSRS (OP) & 49.6\% & 44.1\% & 38.7\% & 35.2\% & 28.1\% & 19.7\% &  &  &  &  &  &  \\
 &  & DSRS(Ours) & \textbf{50.3±0.8\%\textsuperscript{*}} & \multicolumn{1}{l}{\textbf{44.8±0.2\%}\textsuperscript{*}} & \textbf{39.8±0.6\%\textsuperscript{*}} & \multicolumn{1}{l}{\textbf{34.2±0.3\%}} & \textbf{27.7±0.5\%} & \textbf{21.1±0.1\%\textsuperscript{*}} &  &  &  &  &  &  \\ \cline{2-15} 
 & \multirow{3}{*}{SmoothMix} & Neyman-Pearson & 53.2\% & 47.6\% & 40.2\% & 34.2\% & 26.7\% & 19.6\% &  &  &  &  &  &  \\
 &  & DSRS (OP) & 53.3\% & 48.5\% & 42.1\% & 35.9\% & 29.4\% & 21.7\% &  &  &  &  &  &  \\
 &  & DSRS(Ours) & \textbf{52.7±1.6\%} & \textbf{47.4±0.8\%} & \textbf{41.5±1.2\%} & \textbf{35.8±0.8\%} & \textbf{28.8±0.5\%} & \textbf{20.3±0.5\%} &  &  &  &  &  &  \\ \hline
\multirow{9}{*}{1.00} & \multirow{3}{*}{\begin{tabular}[c]{@{}c@{}}Gaussian\\  Augmentation\end{tabular}} & Neyman-Pearson & 40.2\% & 32.6\% & 24.7\% & 18.9\% & 14.9\% & 10.2\% & 7.5\% & 4.1\% & 2.0\% & 0.7\% & 0.1\% & 0.1\% \\
 &  & DSRS (OP) & 40.3\% & 33.1\% & 25.9\% & 20.6\% & 16.1\% & 12.5\% & 8.4\% & 6.4\% & 3.5\% & 1.8\% & 0.7\% & 0.1\% \\
 &  & DSRS(Ours) & \textbf{40.1±1.0\%} & \textbf{32.4±0.5\%} & \textbf{26.1±0.2\%\textsuperscript{*}} & \textbf{20.6±0.4\%} & \textbf{16.5±0.6\%\textsuperscript{*}} & \textbf{12.4±0.4\%} & \textbf{8.9±0.5\%\textsuperscript{*}} & \textbf{6.2±0.3\%} & \textbf{4.1±0.5\%\textsuperscript{*}} & \textbf{2.1±0.3\%\textsuperscript{*}} & \textbf{0.8±0.2\%\textsuperscript{*}} & \textbf{0.2±0.1\%\textsuperscript{*}} \\ \cline{2-15} 
 & \multirow{3}{*}{Consistency} & Neyman-Pearson & 37.2\% & 32.6\% & 29.6\% & 25.9\% & 22.5\% & 19.0\% & 16.4\% & 13.8\% & 11.2\% & 9.0\% & 7.1\% & 5.1\% \\
 &  & DSRS (OP) & 37.1\% & 32.5\% & 29.8\% & 27.1\% & 23.5\% & 20.9\% & 17.6\% & 15.3\% & 13.1\% & 10.9\% & 8.9\% & 6.5\% \\
 &  & DSRS(Ours) & \textbf{36.8±0.2\%} & \textbf{33.1±0.5\%\textsuperscript{*}} & \textbf{29.5±0.9\%} & \textbf{26.2±0.7\%} & \textbf{23.5±0.7\%} & \textbf{20.2±0.1\%} & \textbf{17.5±0.5\%} & \textbf{15.7±0.2\%\textsuperscript{*}} & \textbf{13.4±0.3\%\textsuperscript{*}} & \textbf{11.1±0.3\%\textsuperscript{*}} & \multicolumn{1}{l}{\textbf{8.9±0.1\%}} & \textbf{6.7±0.4\%\textsuperscript{*}} \\ \cline{2-15} 
 & \multirow{3}{*}{SmoothMix} & Neyman-Pearson & 43.2\% & 39.5\% & 33.9\% & 29.1\% & 24.0\% & 20.4\% & 17.0\% & 13.9\% & 10.3\% & 7.8\% & 4.9\% & 2.3\% \\
 &  & DSRS (OP) & 43.2\% & 39.7\% & 34.9\% & 30.0\% & 25.8\% & 22.1\% & 18.7\% & 16.1\% & 13.2\% & 10.2\% & 7.1\% & 3.9\% \\
 &  & DSRS(Ours) & \textbf{43.3±1.8\%\textsuperscript{*}} & \textbf{38.6±1.0\%} & \textbf{34.1±0.9\%} & \textbf{29.6±0.7\%} & \textbf{25.0±0.3\%} & \multicolumn{1}{l}{\textbf{21.8±0.4\%}} & \multicolumn{1}{l}{\textbf{18.3±0.3\%}} & \multicolumn{1}{l}{\textbf{15.3±0.2\%}} & \textbf{13.1±0.4\%} & \textbf{10.2±0.3\%} & \multicolumn{1}{l}{\textbf{7.0±0.5\%}} & \textbf{3.9±0.3\%} \\ \hline
\end{tabular}%
}
\end{table}

\subsection{Results beyond original paper} 
According to the original paper's authors, the DSRS framework provides the best results on the General-Gaussian distributions. In this section, we'll look at the effects of distribution shifts, specifically through Standard Gaussian. Also, we are going to observe the effects of the different training methods as suggested in the main paper, like \textit{PGD(Projected Gradient Descent) Training}\cite{https://doi.org/10.48550/arxiv.1706.06083} and \textit{Interpolated Training}\cite{https://doi.org/10.48550/arxiv.1906.06784} on the Neyman-Pearson and DSRS certified radius and compare these results. Finally, we are going to conduct experiments on various hyperparameters introduced by the certification.
 
\subsubsection{Results on Standard Gaussian}
The following results are based on Standard-Gaussian distribution\cite{https://doi.org/10.48550/arxiv.1902.02918}\cite{https://doi.org/10.48550/arxiv.2002.08118}, and we can see that DSRS performs equivalently to or a bit less than Neyman-Pearson Certification. This is further explained in the ablations for the k hyperparameter.

\begin{table}[H]
\centering
\caption{CIFAR-10:
Standard Gaussian}
\label{table:CIFAR-10 Standard}
\resizebox{16cm}{!}{%
\begin{tabular}{c|c|c|cccccccccccc}
\hline
\multirow{2}{*}{Variance} & \multirow{2}{*}{\begin{tabular}[c]{@{}c@{}}Training\\  Method\end{tabular}} & \multirow{2}{*}{\begin{tabular}[c]{@{}c@{}}Certification\\  Approach\end{tabular}} & \multicolumn{12}{c}{Certified Accuracy under Radius r} \\
 &  &  & 0.25 & 0.50 & 0.75 & 1.00 & 1.25 & 1.50 & 1.75 & 2.00 & 2.25 & 2.50 & 2.75 & 3.00 \\ \hline
\multirow{6}{*}{0.25} & \multirow{2}{*}{\begin{tabular}[c]{@{}c@{}}Gaussian\\  Augmentation\end{tabular}} & Neyman-Pearson & 57.9\% & 40.2\% & 24.2\% &  &  &  &  &  &  &  &  &  \\
 &  & DSRS & 57.9\% & 39.7\% & 23.4\% &  &  &  &  &  &  &  &  &  \\ \cline{2-15} 
 & \multirow{2}{*}{Consistency} & Neyman-Pearson & 63.0\% & 53.9\% & 42.6\% &  &  &  &  &  &  &  &  &  \\
 &  & DSRS & 63.0\% & 53.7\% & 42.4\% &  &  &  &  &  &  &  &  &  \\ \cline{2-15} 
 & \multirow{2}{*}{SmoothMix} & Neyman-Pearson & 61.6\% & 54.5\% & 46.7\% &  &  &  &  &  &  &  &  &  \\
 &  & DSRS & 61.5\% & 54.5\% & 46.2\% &  &  &  &  &  &  &  &  &  \\ \hline
\multirow{6}{*}{0.50} & \multirow{2}{*}{\begin{tabular}[c]{@{}c@{}}Gaussian\\  Augmentation\end{tabular}} & Neyman-Pearson & 51.2\% & 41.2\% & 31.5\% & 20.8\% & 13.3\% & 8.8\% &  &  &  &  &  &  \\
 &  & DSRS & 51.0\% & 41.1\% & 31.4\% & 20.4\% & 13.2\% & 8.5\% &  &  &  &  &  &  \\ \cline{2-15} 
 & \multirow{2}{*}{Consistency} & Neyman-Pearson & 49.8\% & 44.9\% & 39.8\% & 35.1\% & 31.4\% & 25.2\% &  &  &  &  &  &  \\
 &  & DSRS & 49.7\% & 44.7\% & 39.5\% & 35.0\% & 30.1\% & 24.8\% &  &  &  &  &  &  \\ \cline{2-15} 
 & \multirow{2}{*}{SmoothMix} & Neyman-Pearson & 54.5\% & 47.5\% & 41.9\% & 35.8\% & 30.0\% & 25.5\% &  &  &  &  &  &  \\
 &  & DSRS & 54.4\% & 47.5\% & 41.7\% & 35.7\% & 30.0\% & 24.8\% &  &  &  &  &  &  \\ \hline
\multirow{6}{*}{1.00} & \multirow{2}{*}{\begin{tabular}[c]{@{}c@{}}Gaussian\\  Augmentation\end{tabular}} & Neyman-Pearson & 38.4\% & 32.7\% & 25.9\% & 20.8\% & 16.7\% & 12.9\% & 9.3\% & 7.2\% & 4.6\% & 3.2\% & 2.1\% & 1.0\% \\
 &  & DSRS & 38.3\% & 32.4\% & 25.8\% & 20.7\% & 16.7\% & 12.7\% & 9.0\% & 7.0\% & 4.5\% & 3.2\% & 2.0\% & 0.9\% \\ \cline{2-15} 
 & \multirow{2}{*}{Consistency} & Neyman-Pearson & 37.3\% & 33.1\% & 29.8\% & 26.3\% & 23.6\% & 21.0\% & 18.2\% & 16.1\% & 14.5\% & 12.6\% & 10.6\% & 8.9\% \\
 &  & DSRS & 37.1\% & 32.9\% & 29.8\% & 6.3\% & 23.6\% & 20.7\% & 17.9\% & 16.0\% & 14.0\% & 12.4\% & 10.2\% & 8.6\% \\ \cline{2-15} 
 & \multirow{2}{*}{SmoothMix} & Neyman-Pearson & 40.9\% & 36.2\% & 31.9\% & 28.0\% & 24.5\% & 21.5\% & 18.4\% & 15.5\% & 13.8\% & 11.5\% & 9.0\% & 7.4\% \\
 &  & DSRS & 40.9\% & 36.2\% & 31.8\% & 27.8\% & 24.4\% & 21.3\% & 18.3\% & 15.5\% & 13.8\% & 11.3\% & 9.0\% & 7.0\% \\ \hline
\end{tabular}%
}
\end{table}

\subsubsection{Results of adversarial models on General Gaussian and Standard Gaussian}
We then conducted experiments on two, adversarial trained models, i.e., \textit{PGD(Projected Gradient Descent) Training}\cite{https://doi.org/10.48550/arxiv.1706.06083} and \textit{Interpolated Training}\cite{https://doi.org/10.48550/arxiv.1906.06784} trained models and then find the Neyman-Pearson and DSRS-certified radius for both of them.

For the PGD model, we trained Resnet-110\cite{https://doi.org/10.48550/arxiv.1512.03385} on PGD based training and obtained a benign
test accuracy of 82.72\% and adversarial test accuracy of 49.12\% in the initial training.
For Interpolated training, we trained the same for Resnet-110\cite{https://doi.org/10.48550/arxiv.1512.03385} and obtained a benign
test accuracy of 89.55\% and adversarial test accuracy of 47.86\% in the initial training.
Then we trained the same on the \textbf{0.5 noise} using all three training methods, and then
calculated the certification results for both the Standard and General Gaussian distributions .

\begin{table}[H]
\centering
\caption{CIFAR-10:
Adversarially trained Models on General Gaussian}
\label{table:CIFAR-10 Extra General}
\resizebox{16cm}{!}{%
\begin{tabular}{c|c|c|cccccc|c}
\hline
\multirow{3}{*}{\begin{tabular}[c]{@{}c@{}}Training\\  Method\end{tabular}} & \multirow{3}{*}{Model} & \multirow{3}{*}{\begin{tabular}[c]{@{}c@{}}Certification\\  Approach\end{tabular}} & \multicolumn{6}{c|}{\multirow{2}{*}{Certified Accuracy under Radius r}} & \multirow{3}{*}{\begin{tabular}[c]{@{}c@{}}Average\\ Certified \\ Radius\end{tabular}} \\
 &  &  & \multicolumn{6}{c|}{} &  \\
 &  &  & 0.25 & 0.50 & 0.75 & 1.00 & 1.25 & 1.50 &  \\ \hline
\multirow{6}{*}{\begin{tabular}[c]{@{}c@{}}Gaussian\\  Augmentation\end{tabular}} & \multirow{2}{*}{Normal} & Neyman-Pearson & 50.7\% & 38.8\% & 26.8\% & 16.4\% & 8.2\% & 2.7\% & 0.435 \\
 &  & DSRS & 52.8\% & 40.6\% & 29.\% & \multicolumn{1}{l}{20.2\%} & 12.0\% & \multicolumn{1}{l|}{4.7\%} & 0.467 \\ \cline{2-10} 
 & \multirow{2}{*}{Interpolated} & Neyman-Pearson & 59.1\% & 45.4\% & 32.3\% & 20.8\% & 10.5\% & 5.3\% & 0.523 \\
 &  & DSRS & 59.5\% & 47.4\% & 35.1\% & 25.6\% & 15.7\% & 7.4\% & 0.547 \\ \cline{2-10} 
 & \multirow{2}{*}{PGD} & Neyman-Pearson & 58.4\% & 46.4\% & 31.8\% & 20.3\% & 11.0\% & 4.2\% & 0.514 \\
 &  & DSRS & 59.0\% & 47.6\% & 35.2\% & 24.0\% & 14.1\% & 6.4\% & 0.538 \\ \hline
\multirow{6}{*}{Consistency} & \multirow{2}{*}{Normal} & Neyman-Pearson & 49.4\% & 44.4\% & 39.5\% & 31.6\% & 24.8\% & 18.9\% & 0.608 \\
 &  & DSRS & 50.3\% & \multicolumn{1}{l}{44.8\%} & 39.8\% & \multicolumn{1}{l}{34.2\%} & 27.7\% & 21.1\% & 0.610 \\ \cline{2-10} 
 & \multirow{2}{*}{Interpolated} & Neyman-Pearson & 56.4\% & 49.7\% & 42.3\% & 34.4\% & 26.0\% & 18.3\% & 0.659 \\
 &  & DSRS & 56.4\% & 50.5\% & 44.2\% & 37.7\% & 29.4\% & 20.4\% & 0.669 \\ \cline{2-10} 
 & \multirow{2}{*}{PGD} & Neyman-Pearson & 56.0\% & 49.4\% & 42.1\% & 33.8\% & 26.2\% & 17.6\% & 0.652 \\
 &  & DSRS & 56.1\% & 50.3\% & 43.0\% & 36.9\% & 29.8\% & 20.6\% & 0.663 \\ \hline
\multirow{6}{*}{SmoothMix} & \multirow{2}{*}{Normal} & Neyman-Pearson & 53.4\% & 47.3\% & 40.5\% & 33.0\% & 25.3\% & 17.7\% & 0.638 \\
 &  & DSRS & 52.7\% & 47.4\% & 41.5\% & 35.8\% & 28.8\% & 20.3\% & 0.638 \\ \cline{2-10} 
 & \multirow{2}{*}{Interpolated} & Neyman-Pearson & 55.7\% & 48.9\% & 41.3\% & 34.4\% & 27.3\% & 19.1\% & 0.662 \\
 &  & DSRS & 56.3\% & 50.0\% & 43.3\% & 37.4\% & 30.2\% & 21.5\% & 0.670 \\ \cline{2-10} 
 & \multirow{2}{*}{PGD} & Neyman-Pearson & 55.4\% & 48.7\% & 40.6\% & 33.4\% & 26.7\% & 18.7\% & 0.655 \\
 &  & DSRS & 55.6\% & 49.6\% & 42.8\% & 35.8\% & 29.8\% & 21.9\% & 0.661 \\ \hline
\end{tabular}%
}
\end{table}

\begin{table}[H]
\centering
\caption{CIFAR-10:
Adversarially trained Models on Standard Gaussian}
\label{table:CIFAR-10 Extra Standard}
\resizebox{16cm}{!}{%
\begin{tabular}{c|c|c|cccccc|c}
\hline
\multirow{3}{*}{\begin{tabular}[c]{@{}c@{}}Training\\  Method\end{tabular}} & \multirow{3}{*}{Model} & \multirow{3}{*}{\begin{tabular}[c]{@{}c@{}}Certification\\  Approach\end{tabular}} & \multicolumn{6}{c|}{\multirow{2}{*}{Certified Accuracy under Radius r}} & \multirow{3}{*}{\begin{tabular}[c]{@{}c@{}}Average\\ Certified\\ Radius\end{tabular}} \\
 &  &  & \multicolumn{6}{c|}{} &  \\
 &  &  & 0.25 & 0.50 & 0.75 & 1.00 & 1.25 & 1.50 &  \\ \hline
\multirow{6}{*}{\begin{tabular}[c]{@{}c@{}}Gaussian\\  Augmentation\end{tabular}} & \multirow{2}{*}{Normal} & Neyman-Pearson & 51.2\% & 41.2\% & 31.5\% & 20.8\% & 13.3\% & 8.8\% & 0.503 \\
 &  & DSRS & 51.0\% & 41.1\% & 31.4\% & 20.4\% & 13.2\% & 8.5\% & 0.496 \\ \cline{2-10} 
 & \multirow{2}{*}{Interpolated} & Neyman-Pearson & 58.8\% & 47.4\% & 35.6\% & 26.2\% & 17.7\% & 11.2\% & 0.594 \\
 &  & DSRS & 58.6\% & 47.2\% & 35.4\% & 26.2\% & 17.6\% & 10.0\% & 0.583 \\ \cline{2-10} 
 & \multirow{2}{*}{PGD} & Neyman-Pearson & 57.9\% & 47.4\% & 35.3\% & 25.8\% & 16.9\% & 10.8\% & 0.582 \\
 &  & DSRS & 57.7\% & 47.1\% & 35.3\% & 25.6\% & 16.4\% & 10.3\% & 0.571 \\ \hline
\multirow{6}{*}{Consistency} & \multirow{2}{*}{Normal} & Neyman-Pearson & 49.8\% & 44.9\% & 39.8\% & 35.1\% & 30.4\% & 25.2\% & 0.681 \\
 &  & DSRS & 49.7\% & 44.7\% & 39.5\% & 35.0\% & 30.1\% & 24.8\% & 0.658 \\ \cline{2-10} 
 & \multirow{2}{*}{Interpolated} & Neyman-Pearson & 56.4\% & 50.4\% & 44.3\% & 38.3\% & 32.2\% & 25.7\% & 0.743 \\
 &  & DSRS & 56.4\% & 50.3\% & 44.2\% & 38.1\% & 32.0\% & 25.3\% & 0.720 \\ \cline{2-10} 
 & \multirow{2}{*}{PGD} & Neyman-Pearson & 56.0\% & 49.9\% & 43.3\% & 37.5\% & 31.4\% & 25.7\% & 0.735 \\
 &  & DSRS & 55.9\% & 49.9\% & 43.0\% & 37.5\% & 31.3\% & 25.1\% & 0.714 \\ \hline
\multirow{6}{*}{SmoothMix} & \multirow{2}{*}{Normal} & Neyman-Pearson & 54.5\% & 47.5\% & 41.9\% & 35.8\% & 30.0\% & 25.5\% & 0.713 \\
 &  & DSRS & 54.4\% & 47.5\% & 41.7\% & 35.7\% & 30.0\% & 24.8\% & 0.691 \\ \cline{2-10} 
 & \multirow{2}{*}{Interpolated} & Neyman-Pearson & 55.6\% & 50.0\% & 43.8\% & 37.9\% & 32.3\% & 25.9\% & 0.742 \\
 &  & DSRS & 55.5\% & 49.7\% & 43.6\% & 37.8\% & 32.1\% & 25.2\% & 0.719 \\ \cline{2-10} 
 & \multirow{2}{*}{PGD} & Neyman-Pearson & 55.0\% & 50.0\% & 43.1\% & 36.2\% & 31.5\% & 26.6\% & 0.733 \\
 &  & DSRS & 55.0\% & 49.9\% & 43.0\% & 36.1\% & 31.3\% & 25.9\% & 0.710 \\ \hline
\end{tabular}%
}
\end{table}

\indent We found that the results for General Gaussian were far better for DSRS Certification than the Neyman-Pearson but the same can't be seen on Standard Gaussian results. This suggests that the adversarial training methods can influence the robustness radius to a great extent and thus these results bring an interesting future research scope in this area. The results of Standard Gaussian are due to the theoretical limitations of the DSRS framework and can be explained in the ablations of the k hyperparameter, as shown below.

\subsubsection{Effect of ACR on varying hyperparameters N and k}


\begin{figure}
    \centering
    \includegraphics[width = 0.8\textwidth]{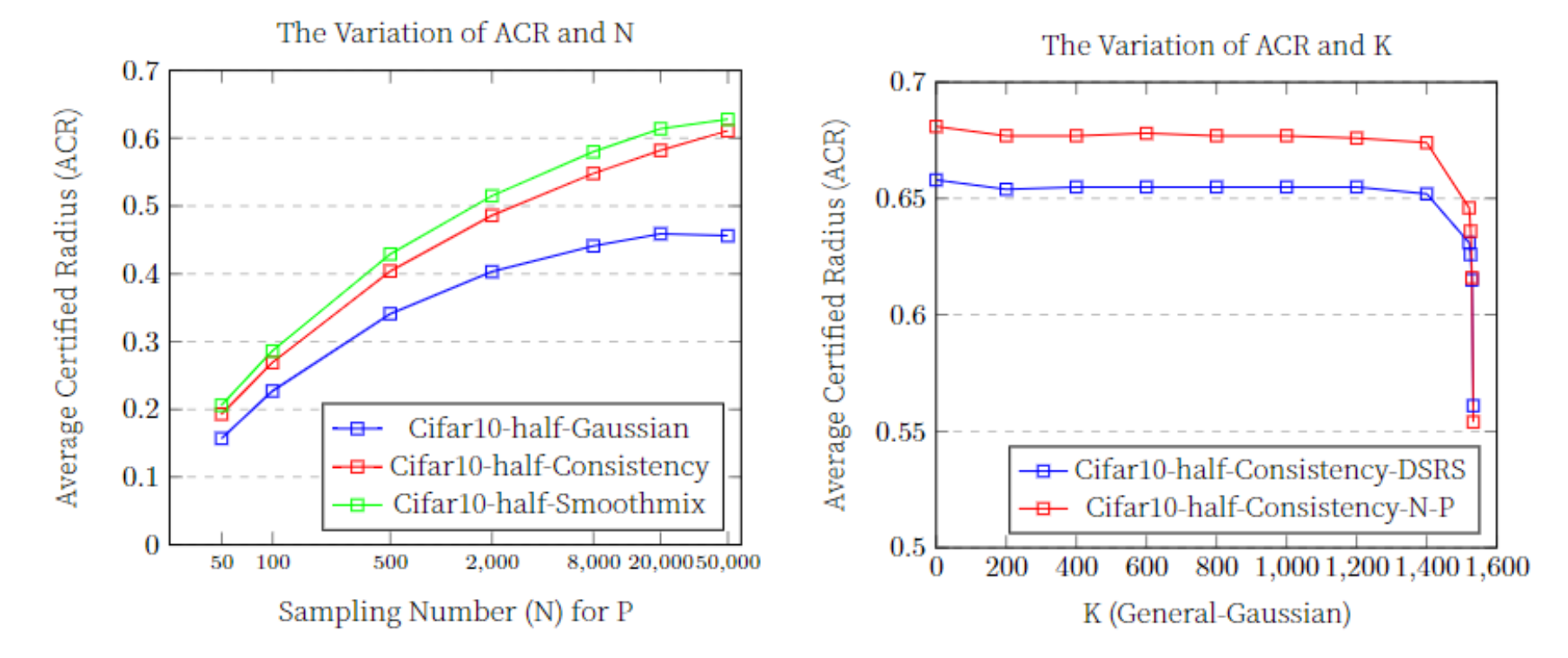}
    \caption{\label{nk} Ablations on the main hyperparameters- N and k shows the insights of the DSRS famework }
    \label{graph:variation in average certified radius wrt nk}
\end{figure}

Here we can see that for N ablations, the results seem to increase linearly for low values of N, i.e., from 50 to 10000, but after that the ACR nearly becomes constant and does not show an appreciable increase. This is in line with similar effects observed on Neyman-Pearson certified radius, but we could have expected DSRS radius to increase more for larger values of N given its theoretical advantages for two sampling distributions.

Also, we can see the ablation on the hyperparameter k which indicates the $\sigma' = \sqrt{\frac{d}{d-2k}}\sigma$ of General Gaussian Distributions\cite{https://doi.org/10.48550/arxiv.2002.09169}. Here, we can see that as we move from $0$ to $d/2-15$ the Neyman-Pearson Certification consistently outperforms the DSRS radius, but it appears to work the other way for the range $d/2-15$ to $d/2$, as theorized by the authors. As a result, the DSRS framework is better suited for higher variance-based gaussian distributions as the parameter k changes the $\sigma'$ which is the variance for generalised gaussian distributions\cite{https://doi.org/10.48550/arxiv.2206.07912}.

\subsubsection{Effect of ACR on varying Sigma P and Sigma Q}
We did experiments to study the distributional shift on $\sigma_P$ and $\sigma_Q$ values. We saw similar results as the authors did with $\sigma_Q=0.4$, $(\sigma_P = 0.5) $, ACR comes out to be maximum.

Also for changing $\sigma_P$ and varying $\sigma_Q$ appropriately, we observed the maximum radius for $\sigma_P=1.0$ which shows a direct relation between higher sigma values of the P and Q distributions and the higher ACR values.

\begin{figure}
    \centering
    \includegraphics[width = 0.8\textwidth]{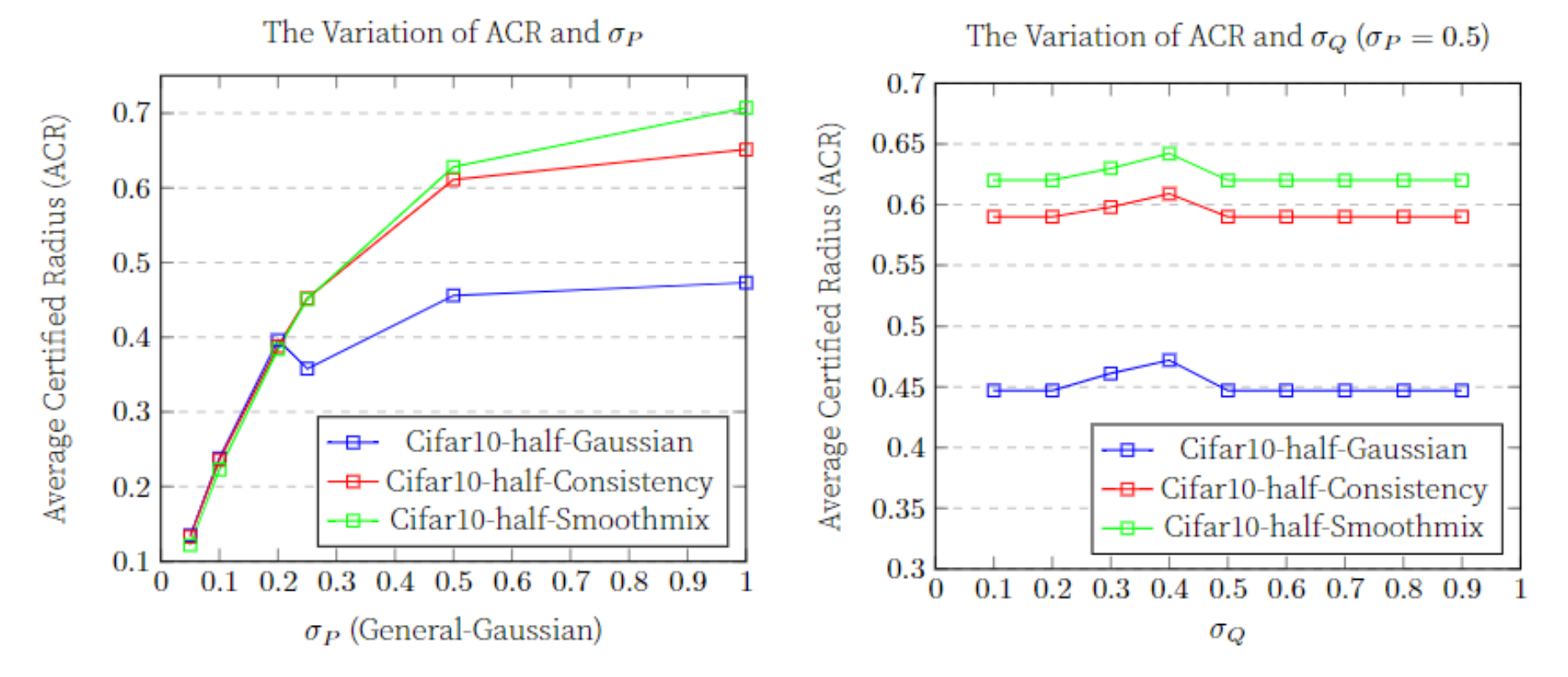}
    \caption{\label{sigma} Variation of ACR for $\sigma$ of P and Q distributions}
    \label{graph:variation in average certified radius wrt pq}
\end{figure}

\section{Discussion}

The results obtained by us largely support the authors' claims that the DSRS verification method is able to provide a tighter robustness certification. While we could not compare and verify the superiority in all cases due to computational limitations, we found the proposed method effective in the various settings we tested. The experiments evaluating the certified radius for various amounts of samplings provide ample evidence that the mechanism can provide comparable results even with much lower sampling numbers, making it much more feasible for real-life deployment.

\subsection{What was easy}
The authors open-sourced the code for the paper. This made it easy to verify many results reported in the paper. The code was clear and concise, allowing us to easily make edits for ablation studies. The mechanism was also mathematically well described in the paper. This helped us conceptually understand the methods. 

\subsection{What was difficult}
The primary difficulty was the excessive requirement of computing capacity for the experiments. Each verification on the MNIST dataset with 50,000 samples took around 10 hours while each verification on the CIFAR-10 dataset took over 25 hours on an NVIDIA V100 GPU. This restricted our selection of datasets and the scope of experiments we were able to conduct. 

\subsection{Recommendations for Reproducibility}
The authors provided very clear instructions for reproducing the main results in the GitHub repository of the paper. They also included the python scripts that can be used for reproducing the main ablations of the paper. They were used to produce the \textit{Main Robustness Verification curve} in Section~\ref{maincurve}. Also, the code was well factored which can make the reproduction effort seemingly easy. The only thing that could have been done better is to provide clear instructions about the ablation scripts as we had a hard time debugging the files for our use.

\subsection{Communication with original authors}\label{AuthorsComment}
We had extensive communication with the original authors throughout the course of the main reproduction of results as well as the ablations. One of the authors, Linyi Li responded well to the PGD\cite{https://doi.org/10.48550/arxiv.1706.06083} and Interpolated training\cite{https://doi.org/10.48550/arxiv.1906.06784} methods we introduced in this paper. Also he helped us with other major ablation experiments and suggested some more experiments, one of which we have included in the final results.

To make the assessment fair and complete, we emailed to the main authors of the paper, particularly Prof. Bo Li and Linyi Li, for their kind feedback on the main report. They gave a very positive feedback for the report and appreciated the ablations we did on the training methods and the hyperparameters. They also acknowledged our efforts and said, "this report, as a role model, is an awesome add-on to our work and greatly benefits the ML community".

\bibliography{bibliography}
\end{document}